\documentclass[11pt]{article}

\usepackage{color}
\usepackage[margin=1in]{geometry}
\usepackage{amssymb}
\usepackage{amsmath}
\usepackage{amsthm}
\usepackage{booktabs}
\usepackage{graphicx}
\usepackage{authblk}

\newtheorem{remark}{Remark}
\newcommand{\R}{\mathbb{R}}

\title{Experimental observation on a low-rank tensor model for eigenvalue problems}
\author{Jun Hu\footnote{School of Mathematical Sciences, Peking University, Beijing 100871, China (hujun@math.pku.edu.cn).}\ \ and\ Pengzhan Jin\footnote{School of Mathematical Sciences, Peking University, Beijing 100871, China (jpz@pku.edu.cn).}
}
\date{}

\begin{document}
	
\maketitle
\begin{abstract}
Here we utilize a low-rank tensor model (LTM) as a function approximator, combined with the gradient descent method, to solve eigenvalue problems including the Laplacian operator and the harmonic oscillator. Experimental results show the superiority of the polynomial-based low-rank tensor model (PLTM) compared to the tensor neural network (TNN). We also test such low-rank architectures for the classification problem on the MNIST dataset.   
\end{abstract}

\section{Introduction}
Neural networks-based machine learning methods are rapidly developed for various numerical problems, such as physics-informed neural networks (PINNs) \cite{lu2021physics,pang2019fpinns,raissi2019physics}, the deep Ritz method \cite{yu2018deep}, and the deep Galerkin method \cite{sirignano2018dgm}. One of the advantages of these approaches is that they show the possibility for solving high-dimensional problems. In \cite{yu2018deep}, the deep learning techniques as well as the Monte-Carlo integration are used to solve eigenvalue problems, which provides a feasible strategy for high-dimensional cases. For the same eigenvalue problems, \cite{wang2022tensor} applies a neural network-based low-rank tensor model, i.e. the tensor neural network (TNN), with a quadrature scheme to perform efficient numerical integration, and thus it achieves a much better result than \cite{yu2018deep}. Furthermore, \cite{wang2022solving} employs the TNN to solve the many-body Schr{\"o}dinger equation, which emerges the practical value of such low-rank approximation method.

Denote by $\phi_{ij}$ the functions defined on compact $\Omega_j$, and then an order-$d$ tensor with respect to these functions can be written as \begin{equation}
u=\sum_{i=1}^r\phi_{i1}\otimes\phi_{i2}\otimes\cdots\otimes\phi_{id},
\end{equation}
whose rank is at most $r$. Consequently, $u$ is defined on $\Omega=\Omega_1\times\Omega_2\times\cdots\times\Omega_d$. Such architecture leads to an efficient high-dimensional integration as
\begin{equation}
\int_{\Omega}u(x)dx=\sum_{i=1}^r\int_{\Omega_1}\phi_{i1}(x_1)dx_1\int_{\Omega_2}\phi_{i2}(x_2)dx_2\cdots\int_{\Omega_d}\phi_{id}(x_d)dx_d.
\end{equation}
This feature is widely used \cite{beylkin2002numerical,litsarev2015fast}, since the computational complexity does not exponentially depend on $d$. Refer to \cite{haastad1989tensor,kolda2009tensor,kruskal1989rank,ryan2002introduction} for more discussions regarding to tensors. As the core idea of the TNN is taking advantage of the low-rank form for high-dimensional integration, we reconsider the necessity of adopting neural networks in such architecture. Therefore, we tend to propose a more concise architecture that only involves analytic basis functions (e.g. polynomials), which allows exact calculations of the integrals. Such low-rank model will greatly reduce the error of numerical integration, even to zero. Below we mainly focus on the performance of the proposed model compared to the TNN, within the same setting of \cite{wang2022tensor}.  

This paper is organized as follows. In Section \ref{sec:ltm}, we introduce the low-rank tensor model. Subsequently, the corresponding method for eigenvalue problem is shown in Section \ref{sec:method}. Section \ref{sec:num} presents several numerical experiments for the proposed method. Finally, Section \ref{sec:con} concludes this work.

\section{Low-rank tensor model}\label{sec:ltm}
The low-rank tensor model (LTM) can be simply written as
\begin{equation}\label{eq:u}
u(x)=\sum_{i=1}^r\prod_{j=1}^d\sum_{k=1}^ba_{ijk}\phi_k(x_j),\quad x=(x_1,\cdots,x_d)\in\Omega\subset\R^d,
\end{equation}
where $r$ is regarded as the rank of this model, $d$ is the dimension of the domain, $\{\phi_k\}_{k=1}^b$ are the chosen basis functions, and $\Omega$ is a compact set in $\R^d$. Here $a_{ijk}$ are learnable parameters.

For multiple outputs case, the model is expressed in the form
\begin{equation}\label{eq:ul}
	u_l(x)=\sum_{i=1}^rw_{li}\prod_{j=1}^d\sum_{k=1}^ba_{ijk}\phi_k(x_j),\quad l=1,2,\cdots,m,
\end{equation}
where $w_{li}$ are also learnable parameters. Then $\mathbf{u}=(u_1,\cdots,u_m)^\top$ maps from $\Omega\subset\R^d$ to $\R^m$.

The approximation property in $H^k$-norm for such model is a direct result from the tensor product of Hilbert spaces $H^k(\Omega)\cong H^k(\Omega_1)\otimes\cdots\otimes H^k(\Omega_d)$ where $\Omega=\Omega_1\times\cdots\times\Omega_d$, $\Omega_i\subset\R$ are compact.
\begin{remark}
If needed, we can also make $b$ and $\phi_k$ depend on $i$ and $j$ as
\begin{equation}
u(x)=\sum_{i=1}^r\prod_{j=1}^d\sum_{k=1}^{b_{ij}}a_{ijk}\phi_{ijk}(x_j).
\end{equation}
For example, a problem with zero boundary and periodic conditions for two different dimensions, is better to adopt polynomial (as in Section \ref{sec:method}) and Fourier bases, respectively.  
\end{remark}

\subsection{Polynomial-based low-rank tensor model}
In this study, we investigate the performance of polynomials as the basis functions. The model is in the form
\begin{equation}\label{eq:u_poly}
	u(x)=\sum_{i=1}^r\prod_{j=1}^d\sum_{k=1}^ba_{ijk}P_k(x_j),\quad x=(x_1,\cdots,x_d)\in\Omega\subset\R^d,
\end{equation}
where $\{P_k\}_{k=1}^b$ are the chosen polynomial bases, such as the Legendre polynomials $L_k$. 

To satisfy required conditions, we may use the linear combination of common basic polynomials as in spectral methods \cite{shen2011spectral}. For example, we can set $P_k=L_{k+1}-L_{k-1}$ to satisfy the zero boundary condition for $\Omega=[-1,1]^d$, which will be tested in our numerical case. For the cuboid domain $\Omega=[s_1,t_1]\times\cdots\times[s_d,t_d]$, $s_j<t_j$, we only need to perform a simple linear transformation, i.e., change $P_k(x_j)$ to $P_k(2\cdot\frac{x_j-s_j}{t_j-s_j}-1)$.

\subsection{Fourier-based low-rank tensor model}
For problems with periodic conditions, we may consider Fourier bases. This case is similar to (\ref{eq:u_poly}), where we employ Fourier basis functions instead of polynomials. This model will be investigated in the future.

\subsection{Tensor neural network}
The tensor neural network (TNN) is studied in \cite{jin2022mionet,wang2022tensor}, which is in the form
\begin{equation}\label{eq:tnn}
u(x)=\sum_{i=1}^r\prod_{j=1}^d\phi_{ij}(x_j),\quad x=(x_1,\cdots,x_d)\in\Omega\subset\R^d,
\end{equation}
where $\Phi_j=(\phi_{1j},\cdots,\phi_{rj})^\top$ mapping from $\R$ to $\R^r$ are modeled as neural networks (e.g. FNNs). The multiple outputs case is written as
\begin{equation}\label{eq:tnnl}
	u_l(x)=\sum_{i=1}^rw_{li}\prod_{j=1}^d\phi_{ij}(x_j),\quad l=1,2,\cdots,m,
\end{equation}
where $w_{li}$ are also learnable parameters. Then $\mathbf{u}=(u_1,\cdots,u_m)^\top$ maps from $\Omega\subset\R^d$ to $\R^m$.

Generally speaking, TNNs replace the basis functions in (\ref{eq:u}) and (\ref{eq:ul}) by neural networks, and these neural networks do not share the parameters with respect to dimension $j$. TNNs will be used as the baseline model in the following experiments.

\subsection{Initialization}
For the polynomial-based low-rank tensor model (PLTM), we set
\begin{equation}\label{eq:init}
	a_{ijk}=\left\{
	\begin{aligned}
		c \quad k=1,\\
		0 \quad k>1,\\
	\end{aligned}
	\right.
\end{equation}
and randomly initialize $w_{li}$, for example, by uniform distribution. Here $c$ is a constant to be tuned, which defaults to 1, if there is no other explanation. The setting of (\ref{eq:init}) is important and necessary to ensure that the products in (\ref{eq:u}) and (\ref{eq:ul}) do not explode or disappear for large $d$.

\section{Method for eigenvalue problem}\label{sec:method}
Here we directly present a typical example to show how to use the previous models to solve the eigenvalue problems.

We are concerned with the following problem:
\begin{equation}\label{eq:eigenvalue}
	\left\{
	\begin{aligned}
		-\Delta u + vu&=\lambda u\quad {\rm in}\quad  \Omega,\\
		u&=0\quad\ \ {\rm on}\quad \partial\Omega,\\
	\end{aligned}
	\right.
\end{equation}
where $\Omega=[s,t]^d,s<t$. The smallest eigenvalue is
\begin{equation}
	\lambda=\min_{0\neq u\in H_0^1(\Omega)}\frac{\int_{\Omega}|\nabla u|^2dx+\int_{\Omega}vu^2dx}{\int_{\Omega}u^2dx}.
\end{equation}
Denote the function space by
\begin{equation}
\mathcal{F}=\left\{u(x)=\sum_{i=1}^r\prod_{j=1}^d\sum_{k=1}^ba_{ijk}P_k(x_j)\Bigg|a_{ijk}\in\R\right\}\subset H_0^1(\Omega),
\end{equation}
where $P_k=L_{k+1}-L_{k-1}$ ($L_k$ are the Legendre polynomials on $[s,t]$), $\theta=\{a_{ijk}\}$ are the learnable parameters, $r$ and $b$ are the hyperparameters to be tuned. Here all the functions in $\mathcal{F}$ satisfy the boundary condition of (\ref{eq:eigenvalue}). Then the loss function can be simply written as
\begin{equation}\label{eq:eigen_loss}
	L(\theta)=L[\tilde{u}]=\frac{\int_{\Omega}|\nabla \tilde{u}|^2dx+\int_{\Omega}v\tilde{u}^2dx}{\int_{\Omega}\tilde{u}^2dx},\quad \tilde{u}\in\mathcal{F},
\end{equation}
where $\theta$ corresponds to the learnable parameters of $\tilde{u}$. Note that all the integrals related to $\tilde{u}$ in (\ref{eq:eigen_loss}) can be exactly calculated due to the low-rank analytic expression of $\tilde{u}$, given any low-rank $v$. For example,

\begin{equation}
\begin{split}
\int_{\Omega}\tilde{u}^2dx&=\int_{\Omega}\left(\sum_{i=1}^r\prod_{j=1}^d\sum_{k=1}^ba_{ijk}P_k(x_j)\right)^2dx \\
&=\int_{\Omega}\sum_{i_1=1}^r\sum_{i_2=1}^r\prod_{j=1}^d\sum_{k_1=1}^b\sum_{k_2=1}^ba_{i_1jk_1}a_{i_2jk_2}P_{k_1}(x_j)P_{k_2}(x_j)dx \\
&=\sum_{i_1=1}^r\sum_{i_2=1}^r\prod_{j=1}^d\sum_{k_1=1}^b\sum_{k_2=1}^ba_{i_1jk_1}a_{i_2jk_2}I_{k_1k_2},
\end{split}
\end{equation}
where
\begin{equation}
I_{k_1k_2}=\int_s^tP_{k_1}(y)P_{k_2}(y)dy.
\end{equation}
The matrix $(I_{ij})\in\R^{b\times b}$ can be easily obtained via the orthogonality of Legendre polynomials. The remaining integrals can be obtained in a similar way.

After initializing $\theta$ as $\theta^{(0)}$, the gradient descent (GD)-based method is applied to optimizing the loss function $L(\theta)$ as
\begin{equation}
\theta^{(k+1)}=\theta^{(k)}-\eta\nabla L(\theta^{(k)}),
\end{equation} 
with a suitable learning rate $\eta$. In the following experiments, we choose Adam \cite{kingma2014adam} as the optimizer. After sufficient training, we obtain the approximate eigenfunction $\tilde{u}^{(K)}$ as well as the corresponding approximate eigenvalue $\lambda^{(K)}=L(\theta^{(K)})$ with a iteration $K$.

\section{Numerical results}\label{sec:num}

\subsection{The Laplacian operator}
Consider the case of the Laplacian operator in Section \ref{sec:method}, where we specify $v=0$ and $[s,t]=[0,1]$, then the exact smallest eigenvalue and eigenfunction are
\begin{equation}
	\lambda=d\pi^2,\quad u(x)=\prod_{i=1}^{d}\sin(\pi x_i).
\end{equation}
The loss function can be written as
\begin{equation}
	L(\theta)=L[\tilde{u}]=\frac{\int_{\Omega}|\nabla \tilde{u}|^2dx}{\int_{\Omega}\tilde{u}^2dx},\quad \tilde{u}\in\mathcal{F}.
\end{equation}
Here we test $d=10$ and $d=512$. The number of bases $b$ and the rank $r$ are both set to 10, and we train this model for 500 iterations with learning rate 0.001 via Adam optimizer. The results are shown in Table \ref{tab:lap}, and they almost achieve the machine precision. The result of the TNN for $d=512$ achieves $1.6\times 10^{-7}$ relative error from \cite{wang2022tensor}.

\begin{table}[htbp]
	\centering
	\begin{tabular}{c|c c}
		\toprule
		$d$ & 10 & 512 \\
		\midrule
		True eigenvalue & $10\pi^2\approx98.69604401089359$ & $512\pi^2\approx5053.237453357751$\\
		Learned eigenvalue & $98.69604401089354$ & $5053.237453358372$ \\
		Relative error & $4.3\times 10^{-16}$ & $1.2\times 10^{-13}$ \\
		\bottomrule
	\end{tabular}
	\caption{Learned eigenvalues of PLTMs for the Laplacian operator. The results almost achieve the machine precision, while the TNN achieves $1.6\times 10^{-7}$ relative error for $d=512$.}
	\label{tab:lap}
\end{table}

\subsection{The harmonic oscillator}
Now we consider the case of the harmonic oscillator, where $v(x)=\sum_{i=1}^d x_i^2$ and $[s,t]=[-5,5]$. Here $v$ is a tensor with a finite rank $d$. The exact smallest eigenvalue and eigenfunction are
\begin{equation}
	\lambda=d,\quad u(x)=\prod_{i=1}^{d}\exp(-\frac{x_i^2}{2}).
\end{equation}
The loss function can be written as
\begin{equation}
	L(\theta)=L[\tilde{u}]=\frac{\int_{\Omega}|\nabla \tilde{u}|^2dx+\sum_{i=1}^d\int_{\Omega}x_i^2\tilde{u}^2dx}{\int_{\Omega}\tilde{u}^2dx},\quad \tilde{u}\in\mathcal{F}.
\end{equation}
We also test $d=10$ and $d=512$. The number of bases $b$ and the rank $r$ are set to 22 and 10, respectively. It is worth noting that we set the initial value $c$ in (\ref{eq:init}) to 0.3 for $d=512$. We train this model for 1000 iterations with learning rate 0.001 via Adam optimizer. The results are shown in Table \ref{tab:har}, and their relative errors are $\sim 10^{-8}$. The result of the TNN for $d=512$ achieves $8.5\times 10^{-6}$ relative error from \cite{wang2022tensor}.

Another noteworthy point is that the loss $L$ will drop down to a value less than the theoretical minimum $\lambda$ if the number of bases $b$ is very large, based on our observation. The machine errors should be carefully taken into account in this model.

\begin{table}[htbp]
	\centering
	\begin{tabular}{c|c c}
		\toprule
		$d$ & 10 & 512 \\
		\midrule
		True eigenvalue & $10$ & $512$\\
		Learned eigenvalue & $10.000000345538101$ & $512.0000063370244$ \\
		Relative error & $3.5\times 10^{-8}$ & $1.2\times 10^{-8}$ \\
		\bottomrule
	\end{tabular}
	\caption{Learned eigenvalues of PLTMs for the harmonic oscillator. Their relative errors are $\sim 10^{-8}$, while the TNN achieves $8.5\times 10^{-6}$ relative error for $d=512$.}
	\label{tab:har}
\end{table}

\subsection{Classification problem}
\begin{figure}[htbp]
	\centering
	\includegraphics[width=0.85\textwidth]{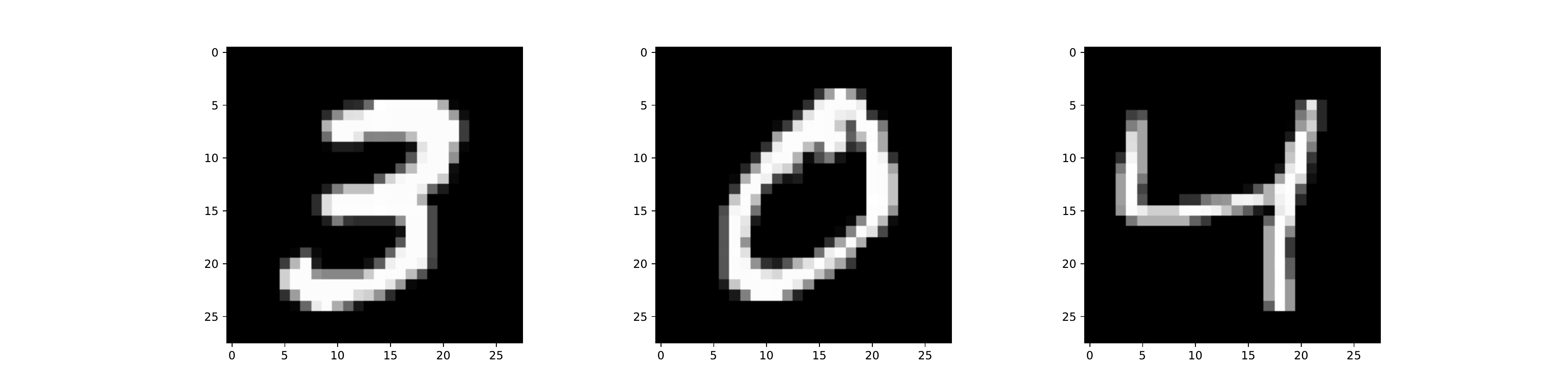}
	\caption{Handwritten digits.}
	\label{fig:digits}
\end{figure}

To investigate the generalization of the LTM, we test the models on the MNIST dataset of handwritten digits \cite{lecun1998gradient}, see Figure \ref{fig:digits} for examples. Each image in MNIST is a $28\times28$ real matrix, which can be regarded as a vector in $\R^{784}$. The images are labeled in 10 categories $\{0,1,\cdots,9\}$, for example, image $x_i$ will be labeled as $y_i=k$ if it represents digit ``$k$''. Denote the dataset by $\{(x_i,y_i)\}_{i=1}^{N}$. 

The classification model can be written as
\begin{equation}
F(x)={\rm Softmax}(\tilde{u}(x)),
\end{equation}
where
\begin{equation}
{\rm Softmax}(x):=\frac{1}{\exp(x_1)+\cdots+\exp(x_m)}\left(\exp(x_1),\cdots,\exp(x_m)\right),\quad x=(x_1,\cdots,x_m),\quad\forall m\in \mathbb{N}^*,
\end{equation}
which is in fact the normalization of the input. Here $\tilde{u}:\R^{784}\to\R^{10}$ is the used function model such as the FNN and the LTM. Denote $F=(F_0,\cdots,F_9)$, then the loss function is
\begin{equation}
L(\theta)=L[\tilde{u}]=-\sum_{i=1}^N\ln(F_{y_i}(x_i)).
\end{equation}
By optimizing $L$, we obtain the $F$ as a classifier. For an unknown input image $x$, we regard the index of the maximum component of $F(x)$ as its prediction. Here we test four models, i.e., the nonlinear FNN (more than one layer), the linear FNN (one linear layer without activation), the PLTM, and the TNN. As for the PLTM, we directly apply Legendre polynomials as the basis functions.
 
The results are shown in Table \ref{tab:mnist}. The nonlinear FNN can achieve accuracy $\approx98\%$, while the linear FNN achieves $\approx92\%$. Both the PLTM and the TNN achieve accuracy $\approx95\%$. In this case, there is no obvious difference between the PLTM and the TNN, and they are both worse than the nonlinear FNN but better than the linear FNN.

\begin{table}[htbp]
	\centering
	\begin{tabular}{c|c c c c}
		\toprule
		 & FNN (nonlinear) & FNN (linear) & PLTM & TNN \\
		\midrule
		MNIST & $\approx98\%$ & $\approx92\%$ & $\approx95\%$ & $\approx95\%$\\
		\bottomrule
	\end{tabular}
	\caption{Different function models for the classification problem on the MNIST dataset.}
	\label{tab:mnist}
\end{table}

\section{Conclusions}\label{sec:con}
We solve the eigenvalue problems including the Laplacian operator and the harmonic oscillator via the low-rank approximation method, where we exploit the polynomial-based low-rank tensor model (PLTM) and the tensor neural network (TNN). There are several points revealing the superiority of the PLTM. (i) Compared to the TNN which requires the numerical integration to approximate the involved integrals, the PLTM allows exact calculations of the integrals. (ii) The results show that the PLTM is more accurate than the TNN. (iii) Moreover, the training process of the PLTM is much faster than the TNN. To investigate the generalization of the PLTM and the TNN, we further test the classification problem on the MNIST dataset, and find that there is no obvious difference between the PLTM and the TNN.

More complicated examples as well as different basis functions (e.g. Fourier basis) are considered in the future.

\bibliographystyle{abbrv}
\bibliography{references}
	
\end{document}